\documentclass[runningheads]{llncs}
\usepackage[T1]{fontenc}
\usepackage{graphicx}
\usepackage{microtype}
\usepackage{hyperref}
\usepackage{url}
\usepackage{booktabs}
\usepackage{tikz}
\usepackage{tcolorbox}
\usepackage{xcolor}
\usepackage{amssymb}
\usepackage{hyperref}
\begin{document}
\title{Assessing and Enhancing the Robustness of Large Language Models with Task Structure Variations for Logical Reasoning}
\author{Qiming Bao\inst{1,2}\orcidID{0000-0002-1000-7383} \and
Gaël Gendron\inst{1}\orcidID{0000-0002-2457-934X} \and
Alex Yuxuan Peng\inst{1}\orcidID{0000-0002-9922-5781} \and
Wanjun Zhong\inst{3}\orcidID{0009-0007-2236-228X} \and
Neset Tan\inst{1}\orcidID{0000-0001-6201-7295} \and
Yang Chen\inst{1}\orcidID{0000-0002-1148-3920} \and
Michael Witbrock\inst{1}\orcidID{0000-0002-7554-0971} \and
Jiamou Liu\inst{1}\orcidID{0000-0002-0824-0899}}
\authorrunning{Bao, Q., et al.}
\institute{Strong AI Lab, NAOInstitute, Waipapa Taumata Rau - The University of Auckland \and
Xtracta, New Zealand \and School of Computer Science and Engineering, Sun Yat-Sen University \\
\email{\{qbao775, ggen187\}@aucklanduni.ac.nz}\\
}
\maketitle              % typeset the header of the contribution
\begin{abstract}
Large language models (LLMs), such as LLaMA, Alpaca, Vicuna, GPT-3.5 and GPT-4, have advanced the performance of AI systems on various natural language processing tasks to human-like levels. %For instance, ChatGPT and GPT-4 demonstrated the abilities to understand and complete conversations in a human-like manner. 
However, their generalisation and robustness when performing logical reasoning has not been sufficiently assessed. To comprehensively evaluate this ability, we develop three new logical reasoning datasets named ``ReClor-plus'', ``LogiQA-plus'' and ``LogiQAv2-plus'' that extend standard logical reasoning datasets to evaluate the robustness of the LLM's reasoning. For each, we create three subsets: the first with randomly shuffled options, the second with the correct choices replaced by ``none of the other options is correct'', and the third with a combination of shuffling and substitution. Experiments on these datasets show that these simple augmentations greatly hinder the models' performance.
% , and fine-tuned language models. 
Despite their high performance on the original publicly available datasets, we find that all models perform poorly on these newly constructed datasets. We also demonstrate that introducing task variations into the training set can markedly improve the model's performance on both the original and our developed datasets.
% We observe that  perturbing a significant percentage (like 50\%) of the data with task variation in the large training set (more than 10,000 samples) like LogiQAv2 can notably enhance the model's generalisation and robustness in logical reasoning tasks. 
Finally, we show that applying logic-driven data augmentation for fine-tuning and prompting can enhance generalisation in both discriminative and generative models, offering a path to improving their robustness for tasks involving logical reasoning. Source code and data are made publicly available at \footnote{\url{https://github.com/Strong-AI-Lab/Logical-and-abstract-reasoning}}.

\end{abstract}
\section{Introduction}
By leveraging the vast amounts of data available on the internet, large language models have achieved great performance on various tasks~\cite{brown2020language,wei2022emergent}. However, training data that is confined to a single task or structure may lead to overfitting to specific tasks, consequently diminishing the model's generalisation capabilities. Specifically, there is a dearth of both quantity and quality of data on logical reasoning available on the internet~\cite{10174688}. This scarcity results in limitations for large language models when tackling complex logical reasoning tasks~\cite{wang-etal-2022-logic}. Consequently, ensuring high quality and diversity in training data becomes essential, as it can significantly aid large language models in enhancing their robustness on reasoning tasks. 
%As a result, our research has emerged as a pivotal challenge in the realm of web data quality.
% Large language models have shown emergent abilities not present in smaller models and have achieved great performance on a wide range of tasks \cite{brown2020language,wei2022emergent}. Emergent abilities range from producing coherent text to answering questions, writing, language translation, poetry generation, and even code writing. This multifaceted capability is not derived from task-specific models, but ``emerges'' as the model learns to predict the next token in a sentence based on its training data. 
% Therefore, We want to know whether large language model can also exhibit good performance on complex logical reasoning tasks. 
Current logical reasoning datasets do not truly represent the reasoning abilities of large language models, as making small modifications significantly degrades performance.  Many logical reasoning datasets are designed to select the precise correct option, models might adapt to choose answers that merely resemble the correct one. Moreover, since numerous public logical reasoning datasets were published prior to the training of these large language models, the models could have been trained using these datasets \cite{DBLP:journals/corr/abs-2312-16337}. Consequently, these models might inadvertently recall the location of the correct answer.

To address the above issues, We propose a data perturbation procedure and apply it to three existing logical reasoning datasets, resulting in the creation of ``ReClor-plus'', ``LogiQA-plus'' and ``LogiQA v2-plus''. These new datasets feature three subsets to evaluate the generalisation and robustness of large language models. These new datasets contain modifications of the task structure of existing logical reasoning datasets. We do not change the semantics of the original context, and only modify its structure (orders, forms). We perform systematic experiments with our logical reasoning datasets to investigate the models' generalisation and robustness on three main aspects. First, we evaluate the in-context generalisation and robustness of large language models on logical reasoning tasks. Second, we perform instruction fine-tuning, instruction prompting and logic-driven data augmentation to evaluate whether those methods can help improve the performance of the models. Third, we investigate how different proportions of data perturbation on the training set can help models improve their generalisation and robustness. 
% The data perturbation we implemented involved perturbing various ratios of the training set, ranging from 5\% to 50\%, by shuffling the order of the options and replacing the correct answer with `none of the other options is correct'. 
Fourth, since model with larger parameter scale demonstrate better performance, it is underexplored how model scale will influence model's performance on logical reasoning. we discuss whether the number of parameters in the model can influence the model's generalisation and robustness. 
% \begin{figure}[t]
% \centering
% \begin{tcolorbox}[title=O.O.D Shuffle and Replace Answer Cases]

% \textbf{Instruction:} Can you predict the correct option for the given input?

% \textbf{Input:} \textbf{Given context}: If you have no keyboarding skills at all, you will not be able to use a computer. And if you are not able to use a computer, you will not be able to write your essays using a word processing program. \textbf{Question}: If the statements above are true, which one of the following must be true? \textbf{Please only return the letter in front of your predict correct option, A, B, C or D.} \textbf{A.} If you are not able to write your essays using a word processing program, you have no keyboarding skills. \textbf{B.} If you are not able to write your essays using a word processing program, you are not able to use a computer. \textbf{C.} \textbf{None of the other options are correct.} \textbf{D.} If you have some keyboarding skills, you will be able to write your essays using a word processing program.

% \textbf{Response:} ``C''
% \end{tcolorbox}
% \caption{Example of out-of-distribution case where we replace the correct answer with \textbf{``none of the other options are correct''} and also shuffle the order of options.}
% \label{fig:ood-shuffle-rep-ans}
% \end{figure}
% We conduct systematic experiments with this dataset to evaluate to what extent large language models can perform out-of-distribution logical reasoning. We aim to improve the current performance of large language models on logical reasoning tasks, especially in out-of-distribution scenarios. 

Our benchmark supports both discriminative large language models like LReasoner~\cite{wang-etal-2022-logic}, MERIt~\cite{jiao2022merit}, and AMR-LDA~\cite{bao-etal-2024-abstract} and generative large language models like GPT-3.5~\cite{openai2023chatgpt}, GPT-4~\cite{openai2023gpt4}, LLaMA~\cite{touvron2023llama}, Alpaca~\cite{alpaca} and Vicuna~\cite{vicuna2023}. 
% The version of GPT-4 that we use does not include multimodal training and its training details are not publicly available.

Our main findings can be summarised as follows:

\begin{itemize}
    \item We find that existing large language models like GPT-3.5 and GPT-4 perform well on logical reasoning tasks in the original format but their performance drops on our new formats, suggesting that the models may have seen these datasets during training and failed to acquire generalised logical reasoning capabilities.

    \item We find that instruction fine-tuning can help large language models increase their generalisation and robustness on logical reasoning tasks. In particular, fine-tuned discriminative large language models often demonstrate permutation invariance. Furthermore, applying logic-driven data augmentation for fine-tuning, combined with prompting, can enhance the generalisation performance of both discriminative large language models and generative large language models.
    
    \item We find that, for large training set sizes (more than 10,000 training samples), high ratio of perturbated data (shuffled and substituted) can help increase generative large language model's performance on most logical reasoning tasks. However, this does not work with small training sets. 

    % \item We find that for large training sets (over 10,000 samples), changing 50\% of the training set to out-of-distribution data improves the performance on both leaked out-of-distribution tasks and original tasks. However, for the smaller training sets, a lower ratio (10-15\%) yields similar enhancements. 
    
    \item Finally, we find surprisingly that there is no direct correlation between the model's size (from LLaMA-7B to LLaMA-65B) and its generalisation and robustness on logical reasoning tasks. Contrary to intuition and observations from other tasks \cite{touvron2023llama}, a larger model does not necessarily guarantee better generalisation and robustness on logical reasoning tasks.
\end{itemize}

\section{Related Work}

Research on the generalisation and robustness of large language models for logical reasoning mainly focuses on synthetic natural language reasoning. Initial findings indicate that transformers can be trained on multi-hop reasoning tasks and substantially generalise to deeper unseen reasoning depths, although it can be challenging on paraphrased synthetic test sets \cite{clark2021transformers}. Efforts to enhance generalisation to deeper multi-step reasoning include the introduction of PARARULE-Plus, providing data augmentation on reasoning depths between 2 to 5 \cite{bao2022multi}. AbductionRules incorporates abductive reasoning to understand and answer the multi-step reasoning task~\cite{young-etal-2022-abductionrules}. ROBUSTLR is a challenging dataset considering conjunction, negation, and utilising logical equivalence for paraphrasing~\cite{sanyal2022robustlr}. 

The existing synthetic multi-step reasoning datasets often lack the complexity and diversity found in real-world data. They might be generated using a limited set of rules or scenarios, which can lead to a narrower scope of logical reasoning challenges. ReClor~\cite{yu2020reclor} and LogiQA~\cite{liu2020logiqa} are challenging reading comprehension datasets derived from real-world examinations such as the GMAT, LSAT, and national civil servant exams \cite{yu2020reclor,liu2020logiqa,liu2023evaluating}. An enhanced version, LogiQAv2, incorporates additional data from Chinese civil servant examinations \cite{liu2023evaluating}. Evaluations reveal that the implementation of discourse order-oriented loss functions, specifically Sentence Order Prediction (SOP) and Next Sentence Prediction (NSP), enhances the performance of models like ALBERT and BERT on reasoning tasks \cite{Lan2020ALBERT:,kenton2019bert,li2022eliteplm}. Further assessments of generative large language models, including GPT-3.5 and GPT-4, on ReClor, LogiQA, and LogiQAv2, have demonstrated their commendable performance. However, it is unclear how robust large language models are in real-world logical reasoning tasks \cite{liu2023evaluating}.

\cite{bao-etal-2024-abstract,jiao2022merit,wang-etal-2022-logic} utilise reading comprehension tasks requiring logical reasoning to perform experiments and evaluate the logical reasoning capabilities in existing large language models. The model needs to predict the answer by understanding the rules within the context and deducing the conclusion, which cannot be directly found through simple pattern matching from the context. This differs fundamentally from typical reading comprehension tasks.
Existing work focuses on reading comprehension through task structure variation, such as shuffling options or replacing the correct one, yet no studies directly utilize task variation to assess logical reasoning. AddSent~\cite{jia2017adversarial} generates misleading text by modifying the question according to certain rules and manually proofreading; AddAny~\cite{jia2017adversarial} automatically searches for misleading texts word by word across various MRC models; AddAnsCtx~\cite{liu2020robust} generates misleading text by removing answer words from answer sentences. In this paper, we define the generalisation and robustness of models in logical reasoning that these models not only need to solve the original question but also address new questions that have been modified through task structure variations, including shuffling the order of options and replacing the correct answer.

\section{Method}
\begin{figure}[t]
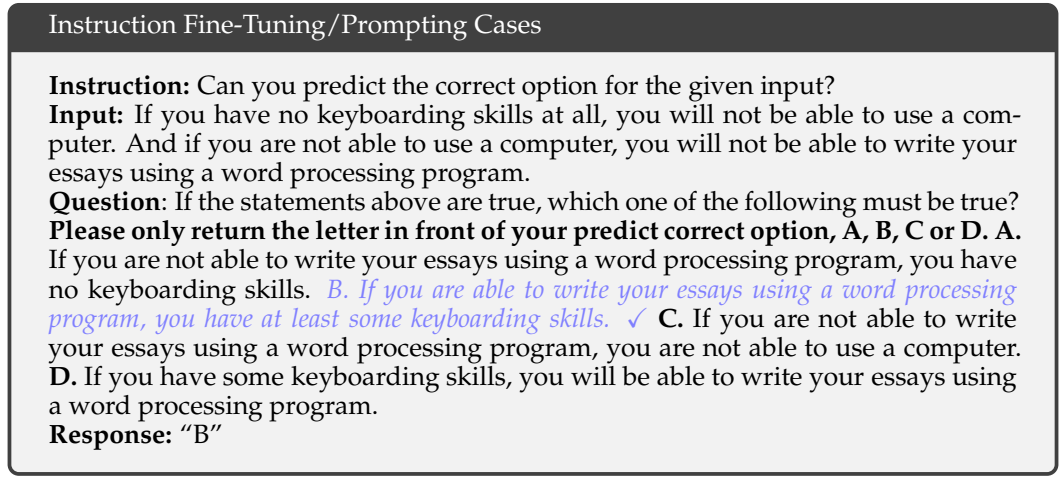

\centering
\begin{tcolorbox}[title=Instruction Fine-Tuning/Prompting Cases]

\textbf{Instruction:} Can you predict the correct option for the given input?

\textbf{Input:} If you have no keyboarding skills at all, you will not be able to use a computer. And if you are not able to use a computer, you will not be able to write your essays using a word processing program. 

\textbf{Question}: If the statements above are true, which one of the following must be true? \textbf{Please only return the letter in front of your predict correct option, A, B, C or D.} \textbf{A.} If you are not able to write your essays using a word processing program, you have no keyboarding skills. \textit{\textcolor{blue!50}{B. If you are able to write your essays using a word processing program, you have at least some keyboarding skills. \checkmark}} \textbf{C.} If you are not able to write your essays using a word processing program, you are not able to use a computer. \textbf{D.} If you have some keyboarding skills, you will be able to write your essays using a word processing program.

\textbf{Response:} ``B''
\end{tcolorbox}
\caption{The instruction fine-tuning involves providing the model with a task description before the input. It includes the Instruction, Input, and Question. The model then gives the expected output. The correct answer is highlighted in blue with a checkmark. Each question has four choices, and only one of them is the correct answer.}
\label{fig:instruction-tuning}
\end{figure}

\begin{figure*}[ht]
  \centering
  \begin{tikzpicture}[node distance=2.75cm,text width=1cm, font=\scriptsize, align=center]
    \node[draw, rectangle, fill=blue!25, text width=1.2cm] (tc) {Generative LLM};
    \node[draw, rectangle, fill=blue!25, text width=1.3cm] (ift) [left of=tc] {Fine-Tuned LLM};
    \node[draw, rectangle, fill=blue!25, text width=1.3cm] (ipt) [right of=tc] {Prompt-Tuned LLM};
    \node[draw, circle, fill=green!25] (mc) [below of=tc] {MCQA};
    \node[draw, circle, fill=green!25] (smc) [below of=ift] {Shuffle MCQA};
    \node[draw, circle, fill=green!25] (rmc) [below of=ipt] {RepAns MCQA};
    \node[draw, rectangle, fill=cyan!25, text width=3.8cm] (enc) [below of=mc] {Fine-Tuned Discriminative LLM}; % New node with increased width

    \draw[->] (tc) -- node[pos=0.5, above]{IFT} (ift);
    \draw[->] (tc) -- node[pos=0.5, above]{IPT} (ipt);
    \draw[->] (mc) -- node[pos=0.5, above]{Shuffle Options} (smc);
    \draw[->] (mc) -- node[pos=0.5, above]{Replace Answer} (rmc);
    \draw[->, dashed,gray!50] (tc) -- node[pos=0.5]{tested on} (mc);
    \draw[->, dashed,gray!50] (ift) -- (mc);
    \draw[->, dashed,gray!50] (ipt) -- (mc);
    \draw[->, dashed,gray!50] (tc) -- (smc);
    \draw[->, dashed,gray!50] (ift) -- node[pos=0.5]{tested on} (smc);
    \draw[->, dashed,gray!50] (ipt) -- (smc);
    \draw[->, dashed,gray!50] (tc) -- (rmc);
    \draw[->, dashed,gray!50] (ift) -- (rmc);
    \draw[->, dashed,gray!50] (ipt) -- node[pos=0.5]{tested on} (rmc);
    
    \draw[->, dashed, gray!50] (enc) -- node[pos=0.5]{tested on} (smc); % New line with 'tested on'
    \draw[->, dashed, gray!50] (enc) -- node[pos=0.5]{tested on} (rmc); % New line with 'tested on'
    \draw[->, dashed, gray!50] (enc) -- node[pos=0.5]{tested on} (mc);  % New line with 'tested on'
\end{tikzpicture}
  \caption{We conduct the IFT and IPT on generative (\textcolor{blue}{blue square}) and discriminative (\textcolor{cyan}{cyan square}) language models, testing them on MCQA datasets (\textcolor{green}{green circles}). These datasets are modified by shuffling options and replacing answers.
  }
  \label{fig:model_data_dep}
\end{figure*}
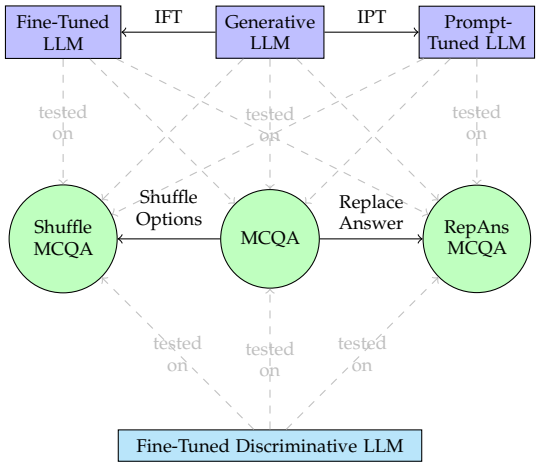

We propose a logical reasoning evaluation benchmark for evaluating the robustness and generalisation of large language models. Figure~\ref{fig:model_data_dep} illustrates the various configurations tested. We propose three \textbf{task structure variations} and apply them to three existing datasets to construct our logical reasoning datasets named ``ReClor-plus'', ``LogiQA-plus'' and ``LogiQAv2-plus''. The task variations are as follows: 1) \textit{\textbf{Shuffle-Order}}: The order of all the options is shuffled. This variation evaluates whether the model is reasoning or remembers the position of the correct answer. 2) \textit{\textbf{Replace-Answer}}: The correct answer is replaced with ``\textit{none of the other options is correct}'' and we add ``\textit{You can also say there is no correct answer}'' at the end of the question. This variation evaluates whether the model understands that apart from the correct option, all other options are incorrect. It is used to detect models returning answers that \textit{look} correct. 3) \textit{\textbf{Shuffle-RepAns}}:  The third split combines the variations from 1) and 2). We apply our variations on ReClor~\cite{yu2020reclor}, LogiQA~\cite{liu2020logiqa} and LogiQAv2~\cite{liu2023evaluating}. The three logical reasoning datasets are all formatted as multiple-choice reading comprehension tasks. An example from ReClor can be found in Figure~\ref{fig:instruction-tuning}. 
% The data statistics for the three datasets can be found in Table~\ref{dataset-description}. 
LogiQA and LogiQAv2 are following the same format as ReClor. For each question in these datasets, there are only four options, and only one of these options is the correct answer. 
% Furthermore, we construct three new logical reasoning datasets named ``ReClor-plus'', ``LogiQA-plus'' and ``LogiQAv2-plus''. For each dataset, we perform three modifications: 1) We randomly shuffle the order of each option. 2) We replace the correct answer by ``none of the other options are correct''. 3) We combine 1) and 2). 
We propose these three datasets to validate the robustness of the model from the following three perspectives: a) We want to evaluate whether the model performs reasoning, instead of just memorizing the position of the correct answer. b) We want to evaluate whether the model understands that, aside from the correct option, all other options are incorrect. We do not want model to simply predict the answer that looks most like the correct answer. c) Building upon the second point, we shuffle the order of the options to judge whether the model can perform more complex reasoning.

% \begin{table}[]
% \centering
% \begin{tabular}{@{}lccc@{}}
% \toprule
% Datasets  & Train & Validation & Test \\ \midrule
% ReClor    & 4638  & 500        & 1000 \\
% LogiQA    & 7376  & 651        & 651  \\
% LogiQAv2 & 12567 & 1569       & 1572 \\ \bottomrule
% \end{tabular}
% \caption{Number of samples in the training, validation, and test set, for ReClor, LogiQA and LogiQAv2.}
% \label{dataset-description}
% \end{table}

Fine-tuned discriminative large language models have to select one answer from the answer set. In constrast, the generative large language models used in next-token prediction setting have to generate a text that matches the correct option letter. To better make a comparison between these two classes of models, we add the following instruction after the question: ``\textbf{Please only return the letter in front of your predict correct option, A, B, C or D.}'' to ease the evaluation. We catch the correct answer using regular expressions from the generated prediction if the prediction does not correspond to the desired format. We use the official Alpaca\footnote{\url{https://github.com/tatsu-lab/stanford_alpaca}\label{alpaca-note}} and Vicuna\footnote{\url{https://github.com/lm-sys/FastChat}} repositories for instruction fine-tuning and instruction prompting. We perform instruction fine-tuning and instruction prompting on both models. For instruction prompting, we use \textit{Chain-of-Thought} prompting to explore how it can help increase the generalisation and robustness of large language models in logical reasoning tasks.

\paragraph{Instruction Fine-Tuning/Prompting} 
Instruction Fine-Tuning (IFT) has been proposed to enhance the performance of large language models on unseen tasks~\cite{mishra-etal-2022-cross,wei2022finetuned}. An example of this approach is depicted in Figure~\ref{fig:instruction-tuning}. An instruction, ``Can you predict the correct option for the given input?'' is added, encompassing the context, question, and each option from the logical reasoning datasets under evaluation. Subsequently, a sentence is introduced: ``Please only return the letter in front of your predicted correct option, A, B, C, or D.'' The process entails training the model on pairs of instructions and corresponding responses. Contrastingly, while instruction prompting employs a similar input format, it is utilised during inference rather than training. Two methods of instruction prompting (IPT) are implemented: a zero-shot evaluation with a format identical to instruction fine-tuning, and a second approach, Chain-of-Thought \cite{wei2022chain}, specifically applied to enhance performance in logical reasoning. The latter integrates the following prompt: ``\textbf{\textit{Describe every step of your reasoning before proposing a solution. When giving the solution, start your sentence with ‘ANSWER:’ }}'' into the instructions to encourage the model to reason.

We also investigate if adding our perturbed sets into the training of the LLMs can help increase performance, and what proportion of perturbed data is required. 
We perform instruction fine-tuning with ReClor, LogiQA and LogiQAv2 separately. We use different ratios of perturbed sets when shuffling and replacing the order of options.

\section{Experiments}\label{sec:experiment}
\subsection{Experiment Setup}
\paragraph{Modeling Choices} We use GPT-3.5-Turbo for the GPT-3.5 experiments and GPT-4 with 8192 tokens for the GPT-4 experiments. For fine-tuned discriminative large language models, we refer to the code from the ReClor leaderboard\footnote{\url{https://github.com/yuweihao/reclor}}. The appendix demonstrates a more detailed hyperparameter setting and model selection.

% This way, the model learns to associate the specific instruction with the appropriate response, making it more efficient and precise when performing the instructed task.

% \paragraph{Task Structure Variations} We specifically target real-world logical reasoning reading comprehension datasets, making variations only in the options. We take into account both the order of the options and the form of the correct option, which can contribute to a more comprehensive exploration of the model's logical reasoning capabilities.
\begin{table*}[]
\centering
\resizebox{1\textwidth}{!}{
\begin{tabular}{@{}lcccccccccccc@{}}
\toprule
Datasets $\rightarrow$       & \multicolumn{4}{c}{ReClor}                                                             & \multicolumn{4}{c}{LogiQA}                                                             & \multicolumn{4}{c}{LogiQAv2}                                                          \\
Models $\downarrow$        & Original & \begin{tabular}[c]{@{}c@{}}Shuffle\\ Order\end{tabular} & \begin{tabular}[c]{@{}c@{}}Replace\\ Answer\end{tabular} & \begin{tabular}[c]{@{}c@{}}Shuffle\\ RepAns\end{tabular} & Original & \begin{tabular}[c]{@{}c@{}}Shuffle\\ Order\end{tabular} & \begin{tabular}[c]{@{}c@{}}Replace\\ Answer\end{tabular} & \begin{tabular}[c]{@{}c@{}}Shuffle\\ RepAns\end{tabular} & Original & \begin{tabular}[c]{@{}c@{}}Shuffle\\ Order\end{tabular} & \begin{tabular}[c]{@{}c@{}}Replace\\ Answer\end{tabular} & \begin{tabular}[c]{@{}c@{}}Shuffle\\ RepAns\end{tabular} \\ \midrule
\multicolumn{13}{c}{\emph{Zero-shot evaluation}} \\
Alpaca-7B &  0.0020        &  0.0060       &  0.0100      &  0.0120                                                        &  0.0122        &  0.0122       & 0.0107       & 0.0121                                                        &  0.0216        &  0.0165       &  0.0095      &  0.0121                                                        \\
Vicuna-7B &  0.0960        &  0.1120       &  0.0740      &  0.0640                                                         &  0.2027        &  0.2135      &  0.1735      &  0.1784                                                       &  0.0834        & 0.0618        &  0.0541      &  0.0121                                                       \\
GPT-3.5 &  0.5702       &  0.5734       & 0.1919      &    0.1847                                                       &  0.3763        &  0.3946      &  0.2449      &   0.2583                                                       & 0.5094         &  0.2695       & 0.2675       &  0.2583                                                      \\
GPT-4 &  0.8735        & 0.8405        &  0.1454     &    0.1312                                                       &  0.4324       &  0.5283     &  0.1007      &    0.1686                                                     &  0.5230        &  0.2616      &  0.1731      &  0.1686                                                        \\\midrule
\multicolumn{13}{c}{\emph{ReClor/LogiQA/LogiQAv2 single training set}} \\
Alpaca-7B-IFT &    0.1680      & 0.5280        &  0.2360      & 0.2720                                                         &   0.1105       &  0.3486       &  0.2841      & 0.2273                                                        &  0.1912        & 0.2122        & 0.3658       & 0.1548                                                         \\
Vicuna-7B-IFT &   0.3040       & 0.1760        & 0.0320       &   0.0420                                                        &   0.2503       &  0.1689       &  0.0706      &  0.1198                                                        &   0.1899        &   0.1746      &  0.1797      &   0.1784                                                       \\
LReasoner &   0.7320       &  0.7100       &  0.2320      &    \textbf{0.3420}                                                       &  0.4147        &  0.4316       &  \textbf{0.5176}      &    \textbf{0.5176}                                                      &  0.5685        &   0.5685      &  \textbf{0.4263}      &   \textbf{0.4263}                                                       \\
MERIt &    0.7960      &   0.7960      &  0.2580      &  0.2460                                                         &   0.3794       &  0.3809       &   0.2657     &  0.2703                                                        &  0.7144        &  0.7144       &   0.1873     &    0.1873                                                      \\
AMR-LDA &   0.8120       &  0.8120       &  \textbf{0.3360}      &  0.3360                                                         &  0.4270        &  0.4301       & 0.1720       & 0.1720                                                         &   0.6985       &  0.6978      & 0.1440       &  0.1440                                                        \\\midrule
\multicolumn{13}{c}{\emph{ReClor + LogiQA + LogiQAv2 merged training set}} \\
Alpaca-7B-IFT &   0.7100       &  0.6560       &   0.1380     &     0.1140                                                      &  0.6651        &   0.4854      &   0.2718     &     0.1351                                                     &   0.6411       &  0.2160       &   0.1956     &    0.1128                                                      \\
Vicuna-7B-IFT &  0.3900        & 0.4040        &  0.1500      & 0.1060                                                         & 0.5453         &  0.3840       &  0.2273      &   0.1490                                                       & 0.4913         & 0.1816        &  0.1708      &  0.1121                                  \\ 
MERIt &     0.9660     &  0.9660       &   0.2440      &  0.2440                                                         &  0.7311        &   0.7342      &  0.2119      &   0.2119                                                       &   0.8655        &  0.8661       &   0.2625     &   0.2625                                                       \\
AMR-LDA &   \textbf{0.9700}       & \textbf{0.9700}        &  0.2900      &  0.2900                                                         &   \textbf{0.7557}        &  \textbf{0.7588}       &  0.2549      &  0.2549                                                        &   \textbf{0.8744}       & \textbf{0.8744}        &  0.3212      &   0.3212                                                       \\\bottomrule
\end{tabular}}
\caption{The experiment compares zero-shot and fine-tuned large language models' logical reasoning across original and merged datasets.}
\label{llm-ood-evaluation}
\end{table*}

\subsection{Result on the Original Datasets}

We summarise our primary findings in Table~\ref{llm-ood-evaluation}. In this subsection, we emphasise results from the original ReClor, LogiQA, and LogiQAv2 datasets. Under the zero-shot evaluation setting, both GPT-3.5 and GPT-4 notably outperform Alpaca-7B and Vicuna-7B across the three datasets, with the latter two models underperforming. Among these, GPT-4 achieves the highest accuracy. We then fine-tuned Alpaca-7B and Vicuna-7B using training data from each individual task and subsequently evaluated them. Notably, all the fine-tuned generative large language models display a marked improvement over their zero-shot evaluation performance.

Simultaneously, we fine-tuned LReasoner, MERIt, and AMR-LDA on their respective training sets and found commendable performance across the board. AMR-LDA's accuracy approached that of GPT-4. To further enhance the diversity and volume of the training data, we merged training sets from ReClor, LogiQA, and LogiQA-v2. Both generative and discriminative large language models exhibited improved results in this setting. The performance of the fine-tuned Alpaca-7B approached GPT-4's, while both MERIt and AMR-LDA surpassed GPT-4. These results underscore the significance of data diversity and volume in logical reasoning tasks.

\subsection{Assessing Models' Robustness on Logical Reasoning Tasks}

% We explore more complex logical reasoning scenarios. We generate two new sets. In the first set, we shuffle the order of options as Figure~\ref{fig:ood-order-shuffling} shown. In the second set, we replace the correct option with \textbf{``none of the other options are correct''}, as Figure~\ref{fig:ood-rep-ans} shown. These datasets allow exploring logical reasoning abilities of language models beyond memorisation. 

\paragraph{Generative Large Language Models} 
We evaluated the performance of GPT-3.5 and GPT-4 on various datasets, notably observing a significant performance drop on the LogiQAv2 dataset, particularly in its Shuffle-Order variant, compared to less pronounced declines on other datasets. Given that GPT-3.5 and GPT-4 were trained before 2023, and considering the publication of ReClor and LogiQA before 2022, it's plausible that these datasets were part of their training data. However, since LogiQAv2 was released after 2023, it was not included, making it a reliable measure of the models' reasoning capabilities. To address potential data leakage concerns with ReClor and LogiQA, we applied instruction fine-tuning using these datasets, both individually and combined with LogiQAv2. This approach significantly enhanced performance across the board. Furthermore, we assessed Alpaca-7B and Vicuna-7B, finding that their zero-shot evaluations were below random chance levels, and fine-tuning failed to yield consistent improvements across datasets. The observed variability in performance, with improvements on shuffled datasets when initial performance was poor and vice versa, suggests a lack of reliable logical reasoning and potential overfitting to the training data, with minor accuracy gains likely attributable to random variance rather than genuine progress.

\paragraph{Discriminative Large Language Models} Fine-tuned discriminative large language models such as LReasoner, MERIt, and AMR-LDA exhibit enhanced generalisation performance on logical reasoning questions. These models demonstrate stable performance on both the original and Shuffle-Order sets; however, they experience a significant decline in performance on the Replace-Answer set and Shuffle-RepAns set. The stability observed in the Shuffle-Order set performance can be attributed to the models' handling of input structure, where the input is formed by concatenating the context, question, and each corresponding option (\textbf{Context + [SEP]' + Question + [SEP]' + Option}), with the output being the label for each concatenation. The special token `[SEP]' is used to separate sentences. Thus, shuffling the option order doesn't create new context, question, and option concatenations, maintaining stable performance and permutation invariance. Despite this, discriminative models do not surpass generative models on the Replace-Answer and Shuffle-RepAns sets, suggesting these models might not engage in complex logical reasoning as effectively. Therefore, the overall performance indicates that large language models, including discriminative ones, may lack strong logical reasoning abilities, as detailed in Table~\ref{llm-ood-evaluation}.

\subsection{Chain-of-Thought (CoT) Prompting} We perform further experiments using Chain-of-Thought prompting~\cite{wei2022chain}. As shown in Table~\ref{chain-of-thought}, in the zero-shot evaluation, all generative large language models do not perform well. The performance of GPT-3.5 and GPT-4 on the Shuffle-RepAns set is nearly equivalent to a random guess, while Alpaca-7B and Vicuna-7B fail the task. Employing the CoT prompting doesn't result in a significant difference in performance for these generative large language models compared to when CoT prompting is not used. Only GPT-4 exhibits systematic improvements in accuracy on the Shuffle-RepAns task. Nonetheless, this task remains challenging for these models, including GPT-4. We have included some case studies in the appendix~\ref{sec:appendix} that illustrate the use of CoT prompting to assist GPT-4 in correctly answering questions with intermediate steps. Overall, since CoT prompting does not offer any explicitly useful information as additional input, and the model has not been trained to respond correctly when given the CoT prompting, it is reasonable to expect that it may not perform well in complex logical reasoning scenarios. Some similar results have been discovered by IDOL\cite{xu-etal-2023-idol} and AMR-LDA\cite{bao-etal-2024-abstract}.

\begin{table*}[h]
\centering
\resizebox{0.6\columnwidth}{!}{
\setlength{\tabcolsep}{10pt}
\begin{tabular}{@{}lccc@{}}
\toprule
Models    & \begin{tabular}[c]{@{}c@{}}ReClor\\ Shuffle\\ RepAns\end{tabular} & \begin{tabular}[c]{@{}c@{}}LogiQA\\ Shuffle\\ RepAns\end{tabular} & \begin{tabular}[c]{@{}c@{}}LogiQAv2\\ Shuffle\\ RepAns\end{tabular} \\ \midrule
\multicolumn{4}{c}{\emph{Zero-shot evaluation}}\\
Alpaca-7B  &   0.0120                 &         0.0230           &    0.0121                                                               \\
Alpaca-7B-CoT &   0.0120                                                                &  0.0337                                                                 &  0.0152                                                                    \\
Vicuna-7B &    0.0640                &        0.1797             &  0.1784                                                               \\
Vicuna-7B-CoT &   0.1320                                                                 &    0.1674                                                               &    0.1593                                                                  \\ 
GPT-3.5 &    \textbf{0.1847}                &  0.2286                  &     \textbf{0.2583}                                                               \\
GPT-3.5-CoT &      0.1088                                                              &   0.1674                                                                & 0.1722                                                                    \\ 
GPT-4 &   0.1312                 &        0.1626            &     0.1686                                                                     \\
GPT-4-CoT &    0.1816                                                                &   \textbf{0.2523}                                                                &   0.2177                                                                   \\ 
\bottomrule
\end{tabular}}
\caption{Comparison between base models and models prompted using CoT prompting.}
\label{chain-of-thought}
\end{table*}

\subsection{Logic-Driven Data Augmentation} 

\begin{table*}[h]
\centering
\resizebox{0.6\columnwidth}{!}{
\setlength{\tabcolsep}{10pt}
\begin{tabular}{@{}lccc@{}}
\toprule
Models    & \begin{tabular}[c]{@{}c@{}}ReClor\\ Shuffle\\ RepAns\end{tabular} & \begin{tabular}[c]{@{}c@{}}LogiQA\\ Shuffle\\ RepAns\end{tabular} & \begin{tabular}[c]{@{}c@{}}LogiQAv2\\ Shuffle\\ RepAns\end{tabular} \\ \midrule
\multicolumn{4}{c}{\emph{Zero-shot evaluation}}\\
Alpaca-7B  &    0.0120                &    0.0121                &    0.0121                                                               \\
GPT-3.5  &   0.1847                 &  0.2583                  &  0.2583                                                                 \\GPT-4  &  0.1312                  &   0.1686                 &  0.1686                                                                 \\\midrule
\multicolumn{4}{c}{\emph{ReClor/LogiQA/LogiQAv2 single training set}}\\
Alpaca-7B-IFT  &  0.2720                  &  0.2273                  &   0.1548                                                                \\
~~~ + AMR-LDA & 0.0440                                                                  & 0.0522                                                                  &   0.0548                                                                   \\ \midrule
\multicolumn{4}{c}{\emph{ReClor + LogiQA + LogiQAv2 merged training set}}\\
Alpaca-7B-IFT  &  0.1140                  &  0.1351                  &   0.1128                                                                \\
~~~ + AMR-LDA & 0.0060                                                                  &   0.0245                                                                &   0.0197                                                                   \\ \midrule
\multicolumn{4}{c}{\emph{Prompt augmentation using AMR-LDA}}\\
Alpaca-7B-IPT-LDA  &   0.0300                 &     0.0368               &   0.0331                                                              \\
Alpaca-7B-IFT-LDA  &   0.4800                 &  0.3686                  &   0.2237                                                              \\
GPT-3.5-IPT-LDA  &     0.3667               &  0.4685                  &      0.4971                                                           \\
GPT-4-IPT-LDA &    \textbf{0.8766}                                                              &        \textbf{0.5510}                                                          &  \textbf{0.7027}   \\
\bottomrule
\end{tabular}}
\caption{Accuracy of evaluated models when adding AMR-LDA's logic-driven augmented data into the training set. We evaluate Alpaca-7B after instruction fine-tuning.}
\label{logic-driven-da}
\end{table*}

\begin{table*}[h]
\centering
\resizebox{1\textwidth}{!}{
\begin{tabular}{@{}lcccccccccccc@{}}
\toprule
Datasets $\rightarrow$       & \multicolumn{4}{c}{ReClor}                                                             & \multicolumn{4}{c}{LogiQA}                                                             & \multicolumn{4}{c}{LogiQAv2}                                                          \\
Perturbation Ratio $\downarrow$        & Original & \begin{tabular}[c]{@{}c@{}}Shuffle\\ Order\end{tabular} & \begin{tabular}[c]{@{}c@{}}Replace\\ Answer\end{tabular} & \begin{tabular}[c]{@{}c@{}}Shuffle\\ RepAns\end{tabular} & Original & \begin{tabular}[c]{@{}c@{}}Shuffle\\ Order\end{tabular} & \begin{tabular}[c]{@{}c@{}}Replace\\ Answer\end{tabular} & \begin{tabular}[c]{@{}c@{}}Shuffle\\ RepAns\end{tabular} & Original & \begin{tabular}[c]{@{}c@{}}Shuffle\\ Order\end{tabular} & \begin{tabular}[c]{@{}c@{}}Replace\\ Answer\end{tabular} & \begin{tabular}[c]{@{}c@{}}Shuffle\\ RepAns\end{tabular} \\ \midrule
\multicolumn{13}{c}{\emph{ReClor/LogiQA/LogiQAv2 single training set with different ratio of data perturbation (Shuffle-RepAns)}} \\
0\% &   0.1680       &  \textbf{0.5280}       &  0.2360      &  0.2720                                                        &  0.1105        &  \textbf{0.3486}       &   0.2841     &     0.2273                                                    &   0.1912       &  \textbf{0.2122}       &  0.3658      &  0.1548                                                        \\
5\% &  0.3340        &  0.3720       &  0.1560      &   0.1720                                                        &  0.1490        &    0.1351     & 0.0998       &   0.0921                                                      &  0.2695        &    0.1516      &  0.1338      &    0.1121                                                      \\
10\% &   \textbf{0.4140}      & 0.4320        & 0.2040      &   0.2380                                                        & \textbf{0.3072}         &  0.2826      &  0.2350      &    0.2442                                                      &  0.2262        &   0.0956      &  0.1963      &    0.1727                                                    \\
15\% &  0.3620        & 0.3860        &  \textbf{0.3060}     &   \textbf{0.3340}                                                        &  0.1904       &   0.2027    &  0.2795      &   0.2319                                                      &  \textbf{0.3537}        &   0.1778     & 0.2001       &     0.1727                                                    \\
50\% &    0.1540      &  0.1400       &  0.1660      &   0.1640                                                       &  0.0430        &  0.0537       &  \textbf{0.6728}      &    \textbf{0.6559}                                                      &  \textbf{0.3537}        &  0.2096       & \textbf{0.7686}       &  \textbf{0.7915}                                  \\ 
\bottomrule
\end{tabular}}
\caption{Accuracy of Alpaca-7B model for transfer learning scenarios and different perturbation ratio applied to the training set. To make a fair comparison, We ensure that the size of each training set is consistent.}
\label{alpaca-ood-ift}
\end{table*}

Since logic-driven data augmentation performs well on logical reasoning tasks for fine-tuned discriminative large language models like LReasoner and AMR-LDA, it is worth considering to apply this method to generative large language models trained on next-token prediction task and see if there is an improvement in our logical reasoning tasks for these models. We use the provided augmented data from the authors of AMR-LDA for ReClor, LogiQA and LogiQAv2 datasets and extend the augmented option information to each option in the training set for ReClor, LogiQA and LogiQAv2. The input format for the augmented data is formed as follows: \textbf{context + question + each option + extended option + extended context}. The term `extended option' refers to the use of AMR-LDA to augment the option based on the logical equivalence laws. `Extended context' means using AMR-LDA to augment the context based on the logical equivalence laws. The output is the ID of each option, which can be either A, B, C, or D. We perform instruction fine-tuning using the input and output formats, along with the same instructions mentioned in the experiment setup. 

Table~\ref{logic-driven-da} shows that logic-driven data augmentation is detrimental to the generalisation and robustness of large language models trained using next-token prediction for logical reasoning tasks. We make the hypothesis that logic-driven data augmentation does not directly map to the task of next-token prediction, which may disturb the training of the model. This hypothesis is corroborated by Table~\ref{logic-driven-da}. When we use individual training sets from ReClor, LogiQA, and LogiQAv2 for separate training and testing, there is an observed improvement compared to models that did not undergo such training. However, performance significantly declines when we utilise AMR-LDA to augment data in the training set. This phenomenon is also evident when the training set is expanded to include ReClor, LogiQA, and LogiQAv2 collectively. This suggests that merely increasing the scale of the training set can offer some benefits in enabling the model to tackle more complex logical reasoning tasks. While, if there is a discrepancy between the distributions of the training and test sets, the potential improvements will be constrained. Additionally, using logic-driven data augmentation in the training set, which might distort its distribution, could further deteriorate performance on the test set.

To enhance the performance of models on more complex logical reasoning tasks, we employ logic-driven data augmentation for generative large language models during instruction fine-tuning and prompting. As shown in Table~\ref{logic-driven-da}, for the experiment was conducted under ``prompt augmentation using AMR-LDA'' area, both the training and test sets were augmented using AMR-LDA. We discovered that two logic-driven data augmentation methods can enhance model's performance on task variation. The first method involves applying logic-driven data augmentation to both the training and test sets; this approach was effective for the fine-tuned Alpaca-7B, but not for the Alpaca-7B that hadn't been fine-tuned. The second method involves applying logic-driven data augmentation solely to the test set; this approach was effective for GPT-3.5 and GPT-4. To implement logic-driven data augmentation on the test set, we followed three steps: \textbf{Step 1}: Augment each option and append the augmented text to the original option. \textbf{Step 2}: If an option can be augmented, we then augment the context and append this augmented context to the already augmented option. \textbf{Step 3}: The augmented prompt will be composed of the following elements: \textbf{context + question + each option + extended option + extended context}.
This strategy helped large language models enhance their performance on our task variation. This result indicates that logic-driven data augmentation offers benefits in two aspects. The first is altering the distribution of both the training and test sets. The second is enhancing the prompt or input information for larger language models like GPT-3.5 and GPT-4, which extracts additional details to large language models, thereby enhancing their efficacy in logical reasoning tasks.

\subsection{Transfer Learning with Data Perturbation}
We perform transfer learning to investigate the extent to which incorporating variations of task structure into the training set can help models improve their performance on logical reasoning tasks. Table \ref{alpaca-ood-ift} shows our results. We carried out instruction fine-tuning on Alpaca-7B using individual training sets from ReClor, LogiQA, and LogiQAv2, each with varying data perturbation ratios, specifically focusing on Shuffle-RepAns. The reason we selected Alpaca-7B is that our main experiment, as shown in Table~\ref{llm-ood-evaluation}, involves both Alpaca-7B and Vicuna-7B. These models demonstrate similar performance in Table~\ref{llm-ood-evaluation}. Under the constraint of limited computational resources, we therefore chose Alpaca-7B for a deeper analysis. We observed that using a large training set of LogiQAv2 for fine-tuning (over 10,000 samples). 50\% of the training set is perturbed to the Shuffle-RepAns format, which improves performance on all original, Shuffle-Order, Replace-Answer, and Shuffle-RepAns format logical reasoning tasks. However, when the amount of training samples used for the fine-tuning is less than 10,000, the model does not demonstrate the above phenomenon. We also observe an intriguing phenomenon: the models' performance on the Shuffle-Order set does not improve as the proportion of the Shuffle-RepAns set in the training set increases. Instead, it declines. Particularly on the ReClor and LogiQA datasets, we can clearly see that the higher the task structure perturbation ratio, the worse the model performs on the Shuffle-Order set. We speculate that the model may overfit the data with altered task structure perturbation ratio. These results also support the claim that existing large language models are not robust when solving logical reasoning tasks. To enhance the robustness of logical reasoning in large language models, employing a more expansive training dataset and a higher perturbation ratio demonstrates beneficial.

\subsection{Influence of the Model Size}
We perform additional experiments to see whether, with the same model architecture, a larger model size will show better generalisation and robustness on logical reasoning tasks. We use different size of LLaMA models to perform the experiments from LLaMA-7B to LLaMA-65B. This choice was made because LLaMA provides model checkpoints for various sizes, ranging from 7B to 70B, which were available during the period of our experiment. We utilised the Shuffle-RepAns sets and perform the zero-shot evaluation. Table~\ref{llama-different-size} shows the results. We do not find a significant difference in the Shuffle-RepAns format logical reasoning task with different sizes of models under zero-shot evaluation, without any instruction fine-tuning. In summary, Table~\ref{llama-different-size} sheds light on the nuanced relationship between model size and performance in logical reasoning tasks. While larger models might be expected to perform better due to their increased capacity, the results presented here suggest that model size does not play a significant role in determining overall performance on robust and complex logical reasoning tasks. It's worth noting that our comparisons are based on different model sizes of the same LLaMA base model. Therefore, the conclusions drawn may not necessarily apply to comparisons involving different base models.

\begin{table*}[h]
\centering
\begin{tabular}{@{}lccc@{}}
\toprule
Models    & \begin{tabular}[c]{@{}c@{}}ReClor\\ Shuffle\\ RepAns\end{tabular} & \begin{tabular}[c]{@{}c@{}}LogiQA\\ Shuffle\\ RepAns\end{tabular} & \begin{tabular}[c]{@{}c@{}}LogiQAv2\\ Shuffle\\ RepAns\end{tabular} \\ \midrule
\multicolumn{4}{c}{\emph{Zero-shot evaluation}}\\
LLaMA-7B  &   \textbf{0.1260}                                                                & 0.1167                                                                  &     0.1128                                                                 \\
LLaMA-13B &                                                              0.0660     &  0.1167                                                                 &     0.1013                                                                 \\
LLaMA-30B &   0.0360                                                                & 0.1290                                                                  &    \textbf{0.1172}                                                                  \\
LLaMA-65B &   0.0720                                                                &   \textbf{0.1397}                                                                &   0.1159                                                                   \\ \bottomrule
\end{tabular}
\caption{Comparison between multiple LLaMA model sizes on logical reasoning tasks with structure variations.}
\label{llama-different-size}
\end{table*}

\section{Conclusion}
% We investigate the generalisation and robustness of large language models on logical reasoning tasks. Our systematic experiments reveal a prevalent limitation in the generalisation and robustness of current models on these tasks. 
% However, we observe that large language models show performance improvements through instruction fine-tuning. We also identify that relying purely on CoT prompting does not significantly assist large language models in robust logical reasoning tasks. Our findings indicate that a training set exceeding 10,000 samples such as LogiQAv2 necessitates substantial data perturbation with a modified task structure (Shuffle-RepAns) to enhance the model’s adaptability, a requirement not observed with smaller datasets. A larger model size does not inherently lead to better generalization and robustness when comparing different sizes under the same LLaMA base model. For fine-tuned discriminative models, we have observed that logic-driven data augmentation is conducive to improved performance. Moreover, these models exhibit permutation invariance, meaning that altering the position of the options does not affect their predictions. This suggests that the models are leveraging more than mere memorization. Furthermore, we apply logic-driven data augmentation to the prompting, which helps larger generative large language models like GPT-3.5 and GPT-4 and fine-tuned Alpaca-7B with logic-driven data augmentation improve their performance in our new logical reasoning tasks.
Our study examines the generalisation and robustness of large language models (LLMs) in logical reasoning tasks, revealing significant limitations. We find that instruction fine-tuning enhances performance, while chain-of-thought (CoT) prompting alone falls short in robust reasoning tasks. Analysis shows that datasets larger than 10,000 samples, such as LogiQAv2, require extensive data perturbation and task structure modifications (Shuffle-RepAns) for improved adaptability—a necessity not seen with smaller datasets. Model size, within the same LLaMA base model framework, does not guarantee better generalisation or robustness. We observe that logic-driven data augmentation benefits fine-tuned discriminative models by improving performance and demonstrating permutation invariance, indicating reliance on more than memorisation. Applying logic-driven data augmentation to prompts further aids larger generative models like GPT-3.5, GPT-4, and fine-tuned Alpaca-7B, enhancing their logical reasoning capabilities in novel tasks.

\bibliographystyle{8379}
\bibliography{8379}
%

% \bibliography{iconip2024_conference}
% \bibliographystyle{iconip2024_conference}





% LNCS Homepage, \url{http://www.springer.com/lncs}, last accessed 2023/10/25
% \end{thebibliography}

\appendix
\section{Appendix}
\label{sec:appendix}

\subsection{Hyperparameter Setting}
By following the public parameter values provided by Stanford Alpaca\textsuperscript{\ref{alpaca-note}} for generative large language models, and the default hyperparameters for each discriminative large language model, including AMR-LDA\footnote{\href{https://github.com/Strong-AI-Lab/Logical-Equivalence-driven-AMR-Data-Augmentation-for-Representation-Learning}{AMR-LDA Official GitHub Page}}, LReasoner\footnote{\url{https://github.com/SiyuanWangw/LReasoner/tree/master}}, and MERIt\footnote{\url{https://github.com/SparkJiao/MERIt}}, we use model weights from their official repositories for logical reasoning tasks. For LReasoner, we use ALBERT-XXLarge-v2~\cite{lan2019albert} as the backbone, and for MERIt and AMR-LDA, we use DeBERTaV2-XXLarge~\cite{he2020deberta}. These models are selected based on recommendations from their original papers~\cite{wang-etal-2022-logic,jiao2022merit}. LReasoner's authors did not release LogiQAv2 weights, so we use weights trained on LogiQA for LogiQAv2 evaluations. For fine-tuning Alpaca and Vicuna, we refer to the official Alpaca GitHub repository~\cite{alpaca}. For GPT-3.5 and GPT-4, we introduce a new instruction for the Shuffle-Order set, shuffling the order of options and requesting the model's response. For the Replace-Answer set, we add the instruction "You can select the option that none of the other options is correct" for more complex scenarios. For the Shuffle-RepAns set, we combine both instructions. All experiments run on 8 NVIDIA A100 GPUs with 80G of VRAM. We evaluate on ReClor, LogiQA, and LogiQAv2 validation sets, using modified sets (Shuffle-Order, Replace-Answer, Shuffle-RepAns) and accuracy as the metric, following methodologies from prior studies~\cite{bao-etal-2024-abstract,jiao2022merit,wang-etal-2022-logic}. For prompting GPT-3.5 and GPT-4, we use OpenAI playground's default hyperparameters.

\end{document}